\documentclass[10pt,journal,compsoc]{IEEEtran}
\ifCLASSOPTIONcompsoc
  \usepackage[nocompress]{cite}
\else
  \usepackage{cite}
\fi
\ifCLASSINFOpdf
\else
\fi

\usepackage{epsfig}
\usepackage{graphicx}
\usepackage{amsmath}
\usepackage{amssymb}

\usepackage{algorithm}
\usepackage{algorithmic}
\usepackage{amsmath}
\usepackage{multirow}
\usepackage{tabularx}
\usepackage{diagbox}
\usepackage{footnote}
\usepackage{threeparttable}
\usepackage{xcolor}
\usepackage[pagebackref=true,breaklinks=true,letterpaper=true,colorlinks,citecolor=citecolor,bookmarks=false]{hyperref}
\definecolor{citecolor}{RGB}{34,139,34}

\usepackage{color}

\newcolumntype{L}[1]{>{\raggedright\let\newline\\\arraybackslash\hspace{0pt}}m{#1}}
\newcolumntype{R}[1]{>{\raggedleft\let\newline\\\arraybackslash\hspace{0pt}}m{#1}}
\newcolumntype{C}[1]{>{\centering\let\newline\\\arraybackslash\hspace{0pt}}m{#1}}

\newcolumntype{x}{>\small c}

\def\eg{\emph{e.g}.} 
\def\ie{\emph{i.e}.}

\def\etal{\emph{et al}.}

\newcommand{\red}[1]{#1}

\begin{document}
%
\title{CCNet: Criss-Cross Attention for Semantic Segmentation}
%
%
%
%

\author{Zilong Huang,
        Xinggang Wang,~\IEEEmembership{Member,~IEEE},
        Yunchao Wei,
        Lichao Huang,
        Humphrey Shi,~\IEEEmembership{Member,~IEEE},
        Wenyu Liu,~\IEEEmembership{Senior Member,~IEEE},
        and Thomas S. Huang,~\IEEEmembership{Life Fellow,~IEEE}
\IEEEcompsocitemizethanks{\IEEEcompsocthanksitem Z. Huang, X. Wang and W. Liu are with the School of Electronic Information and Communications, Huazhong University of Science and Technology, Wuhan 430074, China (e-mail: hzl@hust.edu.cn, xgwang@hust.edu.cn, liuwy@hust.edu.cn).
\IEEEcompsocthanksitem  Y. Wei is with the Centre for Artificial Intelligence, Faculty of Engineering and Information Technology, University of Technology Sydney, Ultimo, NSW 2007, Australia. (e-mail: yunchao.wei@uts.edu.au).
\IEEEcompsocthanksitem L. Huang is with Horizon Robotics.  (e-mail: lichao.huang@horizon.ai).
\IEEEcompsocthanksitem H. Shi is with the University of Oregon and the University of Illinois at Urbana-Champaign. (e-mail: shihonghui3@gmail.com).
\IEEEcompsocthanksitem  T. S. Huang was with the University of Illinois at Urbana-Champaign. (e-mail: t-huang1@illinois.edu).
}
\thanks{Corresponding author: Xinggang Wang. Zilong Huang and Xinggang Wang contributed equally to this work.}}

%
%

\markboth{IEEE Transactions on Pattern Analysis and Machine Intelligence, July~2020}%
{Shell \MakeLowercase{\textit{et al.}}: Bare Demo of IEEEtran.cls for Computer Society Journals}
%



\IEEEtitleabstractindextext{%
\begin{abstract}
Contextual information is vital in visual understanding problems, such as semantic segmentation and object detection. We propose a Criss-Cross Network (CCNet) for obtaining full-image contextual information in a very effective and efficient way. Concretely, for each pixel, a novel criss-cross attention module harvests the contextual information of all the pixels on its criss-cross path. By taking a further recurrent operation, each pixel can finally capture the full-image dependencies. Besides, a category consistent loss is proposed to enforce the criss-cross attention module to produce more discriminative features. Overall, CCNet is with the following merits: 1) GPU memory friendly. Compared with the non-local block, the proposed recurrent criss-cross attention module requires $11\times$ less GPU memory usage. 2) High computational efficiency. The recurrent criss-cross attention significantly reduces FLOPs by about $85\%$ of the non-local block. 3) The state-of-the-art performance. We conduct extensive experiments on semantic segmentation benchmarks including Cityscapes, ADE20K, human parsing benchmark LIP, instance segmentation benchmark COCO, video segmentation benchmark CamVid. In particular, our CCNet achieves the mIoU scores of $81.9\%$, $45.76\%$ and $55.47\%$ on the Cityscapes test set, the ADE20K validation set and the LIP validation set respectively, which are the new state-of-the-art results. The source codes are available at \href{https://github.com/speedinghzl/CCNet}{https://github.com/speedinghzl/CCNet}.
\end{abstract}

}

\maketitle

\IEEEdisplaynontitleabstractindextext

%
\IEEEpeerreviewmaketitle

\IEEEraisesectionheading{\section{Introduction}\label{sec:introduction}}

%
%
%
%

\IEEEPARstart{S}{emantic} segmentation, which is a fundamental problem in the computer vision community, aims at assigning semantic class labels to each pixel in a given image. It has been extensively and actively studied in many recent works and is also critical for various significant applications such as autonomous driving~\cite{fritsch2013new}, augmented reality~\cite{azuma1997survey},image editing~\cite{evening2012adobe}, civil engineering~\cite{song2020deep}, remote sensing imagery~\cite{Zheng_2020_CVPR} and agricultural pattern analysis~\cite{chiu2020agriculture,tik20201st}. Specifically, current state-of-the-art semantic segmentation approaches based on the fully convolutional network (FCN)~\cite{long2015fully} have made remarkable progress. However, due to the fixed geometric structures, the conventional FCN is inherently limited to local receptive fields that only provide short-range contextual information. The limitation of insufficient contextual information imposes a great adverse effect on its segmentation accuracy.

\begin{figure}[!t]
    \centering
    \includegraphics[width=0.9\linewidth]{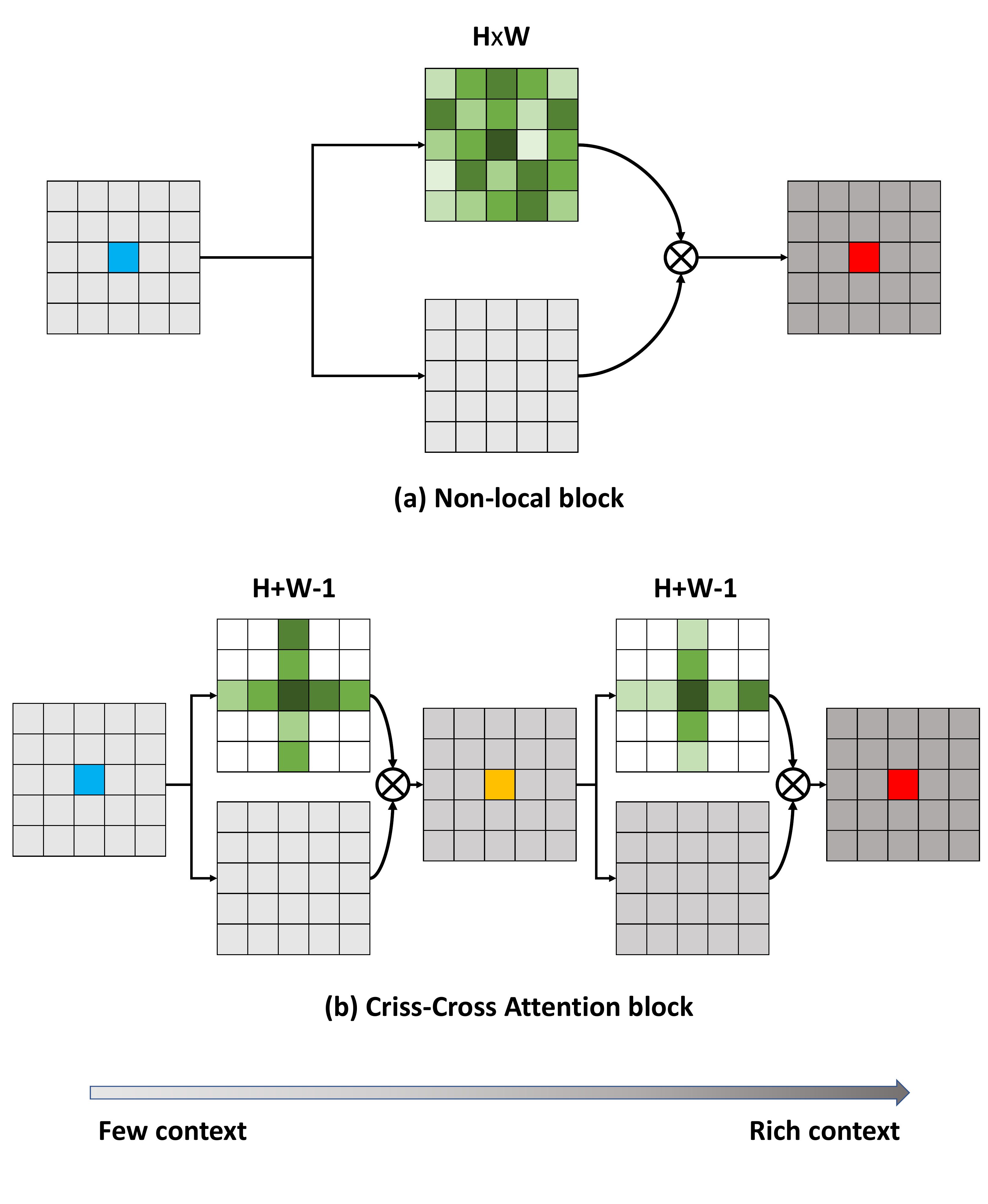}
    \caption{Diagrams of two attention-based context aggregation methods. (a) For each position (\eg, blue), the Non-local module \cite{wang2018non} generates a dense attention map which has $N$ weights (in green). (b) For each position (\eg, blue), the criss-cross attention module generates a sparse attention map which only has about $2\sqrt{N}$ weights. After the recurrent operation, each position (\eg, red) in the final output feature maps can collect information from all pixels. For clear display, residual connections are ignored.}
    \label{fig:motivation}
    \vspace{-5mm}
\end{figure}

To make up for the above deficiency of FCN, some works have been proposed to introduce useful contextual information to benefit the semantic segmentation task. Specifically, Chen \etal~\cite{chen2018deeplab} proposed atrous spatial pyramid pooling module with multi-scale dilation convolutions for contextual information aggregation. Zhao \etal~\cite{zhao2017pyramid} further introduced PSPNet with pyramid pooling module to capture contextual information.
However, the dilated convolution based methods~\cite{chen2017rethinking, chen2018deeplab, Ding_2018_CVPR} collect information from a few surrounding pixels and cannot generate dense contextual information actually. Meanwhile, the pooling based methods~\cite{zhao2017pyramid, zhang2018context} aggregate contextual information in a non-adaptive manner and the homogeneous context extraction procedure is adopted by all image pixels, which does not satisfy the requirement that different pixels need different contextual dependencies.

To incorporate dense and pixel-wise contextual information, some fully-connected graph neural network (GNN)~\cite{scarselli2008graph} methods were proposed to augments traditional convolutional features with an estimated full-image context representation. 
PSANet~\cite{zhao2018psanet} learns to aggregate contextual information for each position via a predicted attention map. Non-local Networks~\cite{wang2018non} utilizes a self-attention mechanism~\cite{cheng2016long, vaswani2017attention}, which enables a single feature from any position to perceive features of all the other positions, thus harvesting full-image contextual information, see Fig.~\ref{fig:motivation}~(a). These non-local operations could be viewed as a densely-connected GNN module based on attention mechanism~\cite{vaswani2017attention}. This feature augmentation method allows a flexible way to represent non-local relations between features and has led to significant improvements in several vision recognition tasks. However, these GNN-based non-local neural networks need to generate huge attention maps to measure the relationships for each pixel-pair, leading to a very high complexity of $\mathcal{O}(N^2)$ for both time and space, where $N$ is the number of input features. Since the dense prediction tasks, such as semantic segmentation, inherently require high resolution feature maps, the non-local based methods will often with high computation complexity and occupy a huge number of GPU memory. Thus, is there an alternative solution to achieve such a target in a more efficient way?


To address the above mentioned issue, our motivation is to replace the common single densely-connected graph with several consecutive sparsely-connected graphs, which usually require much lower computational resources. Without loss of generality, we use two consecutive criss-cross attention modules, in which each one only has sparse connections (about $\sqrt{N}$) for each position in the feature map. For each pixel/position, the criss-cross attention module aggregates contextual information in its horizontal and vertical directions. By serially stacking two criss-cross attention modules, each position can collect contextual information from all pixels in the given image. The above decomposition strategy will greatly reduce the complexities of both time and space from $\mathcal{O}(N^2)$ to $\mathcal{O}(N\sqrt{N})$.

We compare the differences between the non-local module~\cite{wang2018non} and our criss-cross attention module in Fig.~\ref{fig:motivation}. Concretely, both non-local module and criss-cross attention module feed the input feature map to generate an attention map for each position and transform the input feature map into an adapted feature map. Then, a weighted sum is adopted to collecting contextual information from other positions in the adapted feature map based on the attention maps. Different from the dense connections adopted by the non-local module, each position (\eg, blue) in the feature map is sparsely connected with other ones which are in the same row and the same column in our criss-cross attention module, leading to the predicted attention map only has about $2\sqrt{N}$ weights rather than $N$ in non-local module.

To achieve the goal of capturing the full-image dependencies, we innovatively and simply take a recurrent operation for the criss-cross attention module. In particular, the local features are firstly passed through one criss-cross attention module to collect the contextual information in horizontal and vertical directions. Then, by feeding the feature map produced by the first criss-cross attention module into the second one, the additional contextual information obtained from the criss-cross path finally enables the full-image dependencies for all positions. As demonstrated in Fig.~\ref{fig:motivation} (b), each position (\eg red) in the second feature map can collect information from all others to augment the position-wise representations. We share parameters of the criss-cross modules to keep our model slim. Since the input and output are both convolutional feature maps, criss-cross attention module can be easily plugged into any fully convolutional neural network, named as CCNet, for learning  full-image contextual information in an end-to-end manner. Thanks to the good usability of criss-cross attention module, CCNet is straight forward to extend to 3D networks for capturing long-range temporal context information. 

In addition, to drive the proposed recurrent criss-cross attention method to learn more discriminative features, we introduce a category consistent loss to augment CCNet. Particularly, the category consistent loss enforces the network to map each pixel in the image to an n-dimensional vector in the feature space, such that feature vectors of pixels that belong to the same category lie close together while feature vectors of pixels that belong to different categories lie far apart.

We have carried out extensive experiments on multiple large-scale datasets. Our proposed CCNet achieves top performance on four most competitive semantic segmentation datasets, \ie, Cityscapes~\cite{cordts2016cityscapes}, ADE20K~\cite{zhou2017scene}, LIP~\cite{liang2018look} and CamVid~\cite{brostow2009semantic}. In addition, the proposed criss-cross attention even improves the state-of-the-art instance segmentation method, \ie, Mask R-CNN with ResNet-101~\cite{he2016deep}. These results well demonstrate that our criss-cross attention module is generally beneficial to the dense prediction tasks. In summary, our main contributions are three-fold:
\begin{itemize}
     \item We propose a novel criss-cross attention module in this work, which can be leveraged to capture contextual information from full-image dependencies in a more efficient and effective way.
     \item We propose category consistent loss which can enforce  criss-cross attention module to produce more discriminative features.
     \item We propose CCNet by taking advantages of recurrent criss-cross attention module, achieving leading performance on segmentation-based benchmarks, including Cityscapes, ADE20K, LIP, CamVid and COCO.
\end{itemize}

Compare with our original conference version~\cite{huang2018ccnet}, the following improvements are conducted: 1) We further enhance the segmentation ability of CCNet by augmenting a simple yet effective category consistent loss; 2) we propose a more generic CCNet by extending the criss-cross attention module from 2D to 3D; 3) we include more extensive experiments on the LIP, CamVid and COCO datasets to verify the effectiveness and generalization ability of our CCNet.

The rest of this paper is organized as follows. We first
review related work in Section~\ref{Related work} and describe the architecture of our network in Section~\ref{CC network}. In Section~\ref{Experiments}, ablation studies are given and experimental results are analyzed. Section~\ref{Conclusion} presents our conclusion and future work.

\begin{figure*}[!t]
    \centering
    \includegraphics[width=1.0\linewidth]{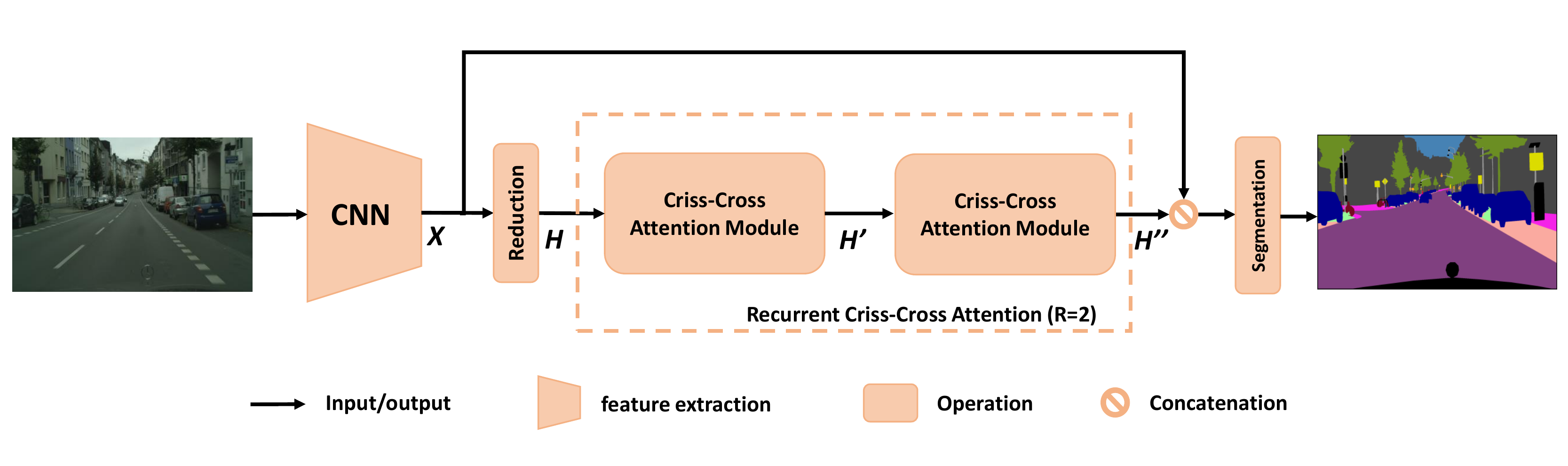}
    \caption{Overview of the proposed CCNet for semantic segmentation.
    }
    \label{fig:architecture}
\end{figure*}

\section{Related work} \label{Related work}

\subsection{Semantic segmentation} 

The last years have seen a renewal of interest on semantic segmentation. FCN~\cite{long2015fully} is the first approach to adopt fully convolutional network for semantic segmentation. Later, FCN-based methods have made remarkable progress in image semantic segmentation. Chen \etal~\cite{chen2014semantic} and Yu \etal~\cite{yu2015multi} removed the last two downsample layers to obtain dense prediction and utilized dilated convolutions to enlarge the receptive field. Unet~\cite{ronneberger2015u}, DeepLabv3+~\cite{chen2018encoder}, MSCI~\cite{lin2018multi}, SPGNet~\cite{chen2019spgnet}, RefineNet~\cite{lin2017refinenet} and DFN~\cite{yu2018learning} adopted encoder-decoder structures that fuse the information in low-level and high-level layers to make dense predictions. The scale-adaptive convolutions (SAC)~\cite{zhang2017scale} and deformable convolutional networks (DCN)~\cite{dai2017deformable} methods improved the standard convolutional operator to handle the deformation and various scales of objects. CRF-RNN~\cite{yu2015multi} and DPN~\cite{liu2015semantic} used Graph model, \ie, CRF, MRF, for semantic segmentation. AAF~\cite{ke2018adaptive} used adversarial learning to capture and match the semantic relations between neighboring pixels in the label space. BiSeNet~\cite{yu2018bisenet} was designed for real-time semantic segmentation. DenseDecoder~\cite{bilinski2018dense} built feature-level long-range skip connections on cascaded architecture. VideoGCRF~\cite{chandra2018deep} used a densely-connected spatio-temporal graph for video semantic segmentation. RTA~\cite{huang2018efficient} proposed the region-based temporal aggregation for leveraging the temporal information in videos. In addition, some works focus on human parsing task. JPPNet~\cite{liang2018look} embed pose estimation into human parsing task. CE2P~\cite{ruan2019devil} proposed a simple yet effective framework for computing context embedding while preserving edges. SANet~\cite{huang2019sanet} used parallel branches with scale attention to handle large scale variance in human parsing. 
Semantic segmentation is also actively studied in the context of domain adaptation and dstillation~\cite{wang2020differential, wang2020alleviating, jiao2019geometry} and weakly supervised setting~\cite{huang2018weakly,wei2018revisiting,qian2019weakly}, etc.

\subsection{Contextual information aggregation}

It is a common practice to aggregate contextual information to augment the feature representation in semantic segmentation networks.
Deeplabv2~\cite{chen2018deeplab} proposed atrous spatial pyramid pooling (ASPP) to use different dilation convolutions to capture contextual information. DenseASPP~\cite{yang2018denseaspp} brought dense connections into ASPP to generate features with various scale. DPC~\cite{chen2018searching} utilized architecture search techniques to build multi-scale architectures for semantic segmentation. Chen \etal~\cite{chen2016attention} made use of several attention masks to fuse feature maps or prediction maps from different branches. PSPNet~\cite{zhao2017pyramid} utilized pyramid spatial pooling to aggregate contextual information. Recently, Zhao \etal~\cite{zhao2018psanet} proposed the point-wise spatial attention network which uses predicted attention map to guide contextual information collection. Auto-Deeplab~\cite{liu2019auto} utilized neural architecture search to search an effective context modeling. He \etal\cite{he2019adaptive} proposed an adaptive pyramid context module for semantic segmentation.
Liu \etal~\cite{liu2017learning} utilized recurrent neural networks (RNNs) to capture long-range dependencies.

There are some works use graph models to model the contextual information. Conditional random field (CRF)~\cite{chen2014semantic, zheng2015conditional, huang2018efficient}, Markov random field (MRF)~\cite{liu2015semantic} were also utilized to capture long-range dependencies for semantic segmentation. 
Vaswani \etal~\cite{vaswani2017attention} applied a self-attention model on machine translation. Wang \etal~\cite{wang2018non} proposed the non-local module to generate the huge attention map by calculating the correlation matrix between each spatial point on the feature maps, then the attention map guided dense contextual information aggregation. OCNet~\cite{yuan2018ocnet} and DANet~\cite{fu2018dual} utilized Non-local module~\cite{wang2018non} to harvest the contextual information. PSA~\cite{zhao2018psanet} learned an attention map to aggregate contextual information for each individual point adaptively and specifically. Chen \etal~\cite{chen2019graph} proposed graph-based global reasoning networks which implements relation reasoning via graph convolution on a small graph.

\textbf{CCNet vs. Non-Local vs. GCN}.
Here, we specifically discuss the differences among GCN~\cite{peng2017large}, Non-local Network~\cite{wang2018non} and CCNet.
In term of contextual information aggregation, only the center point can perceive the contextual information from all pixels by the global convolution filters in GCN~\cite{peng2017large}. In contrast, Non-local Network~\cite{wang2018non} and CCNet guarantee that a pixel at any position perceives contextual information from all pixels. Though GCN~\cite{peng2017large} alternatively decomposes the square-shape convolutional operation to horizontal and vertical linear convolutional operations which is related to CCNet, CCNet takes the criss-cross way to harvest contextual information which is more effective than the horizontal-vertical separate way. Moreover, CCNet is proposed to mimic Non-local Network~\cite{wang2018non} for obtaining dense contextual information through a more effective and efficient recurrent criss-cross attention module, in which dissimilar features get low attention weights and features with high attention weights are similar ones. GCN~\cite{peng2017large} is a conventional convolution neural network, while CCNet is a graph neural network in which each pixel in the convolutional feature map is considered as a node and the relation/context among nodes can be utilized to generate better node features.

\subsection{Graph neural networks}

Our work is related to deep graph neural network (GNN). Prior to graph neural networks, graphical models, such as the conditional random field (CRF)~\cite{chen2014semantic, zheng2015conditional, huang2018efficient}, markov random field (MRF)~\cite{liu2015semantic}, were widely used to model the long-range dependencies for image understanding. GNNs were early studied in \cite{sperduti1997supervised, gori2005new, scarselli2008graph}. Inspired by the success of CNNs, a large number of methods adapt graph structure into CNNs. These methods could be divided into two main steams, the spectral-based approaches~\cite{henaff2015deep, defferrard2016convolutional, kipf2016semi, levie2018cayleynets} and the spatial-based approaches~\cite{atwood2016diffusion, niepert2016learning, gilmer2017neural, wang2018non}. The proposed CCNet belongs to the latter.

\section{Approach} \label{CC network}
In this section, we give the details of the proposed Criss-Cross Network (CCNet) for semantic segmentation. We first present a general framework of our CCNet. Then, the 2D criss-cross attention module which captures contextual information in horizontal and vertical directions will be introduced. To capture the dense and global contextual information, we propose to adopt a recurrent operation for the criss-cross attention module. To further improve RCCA, we introduce a discriminative loss function to drive RCCA to learn category consistent features. Finally we propose the 3D criss-cross attention module for leveraging temporal and spatial contextual information simultaneously.

\subsection{Network Architecture}

The network architecture is given in Fig.~\ref{fig:architecture}.
An input image is passed through a deep convolutional neural network (DCNN), which is designed in a fully convolutional fashion~\cite{chen2018deeplab}, to produce feature map $\mathbf{X}$ with the spatial size of $H \times W$. In order to retain more details and efficiently produce dense feature maps, we remove the last two down-sampling operations and employ dilation convolutions in the subsequent convolutional layers, leading to enlarging the width/height of the output feature map $\mathbf{X}$ to 1/8 of the input image.

Given $\mathbf{X}$, we first apply a convolutional layer to obtain the feature map $\mathbf{H}$ of dimension reduction. Then, $\mathbf{H}$ is fed into the criss-cross attention module to generate a new feature map $\mathbf{H'}$ which aggregate contextual information together for each pixel in its criss-cross path.  The feature map $\mathbf{H'}$ only contains the contextual information in horizontal and vertical directions which are not powerful enough for accurate semantic segmentation. To obtain richer and denser context information, we feed the feature map $\mathbf{H'}$ into the criss-cross attention module again and output the feature map $\mathbf{H''}$. Thus, each position in $\mathbf{H''}$ actually gathers the information from all pixels. Two criss-cross attention modules before and after share the same parameters to avoid adding too many extra parameters. We name this recurrent structure as recurrent criss-cross attention (RCCA) module.

Then, we concatenate the dense contextual feature $\mathbf{H''}$ with the local representation feature $\mathbf{X}$. It is followed by one or several convolutional layers with batch normalization and activation for feature fusion. Finally, the fused features are fed into the segmentation layer to predict the final segmentation result.

\subsection{Criss-Cross Attention} 
\begin{figure}[!t]
    \centering
    \includegraphics[width=1.0\linewidth]{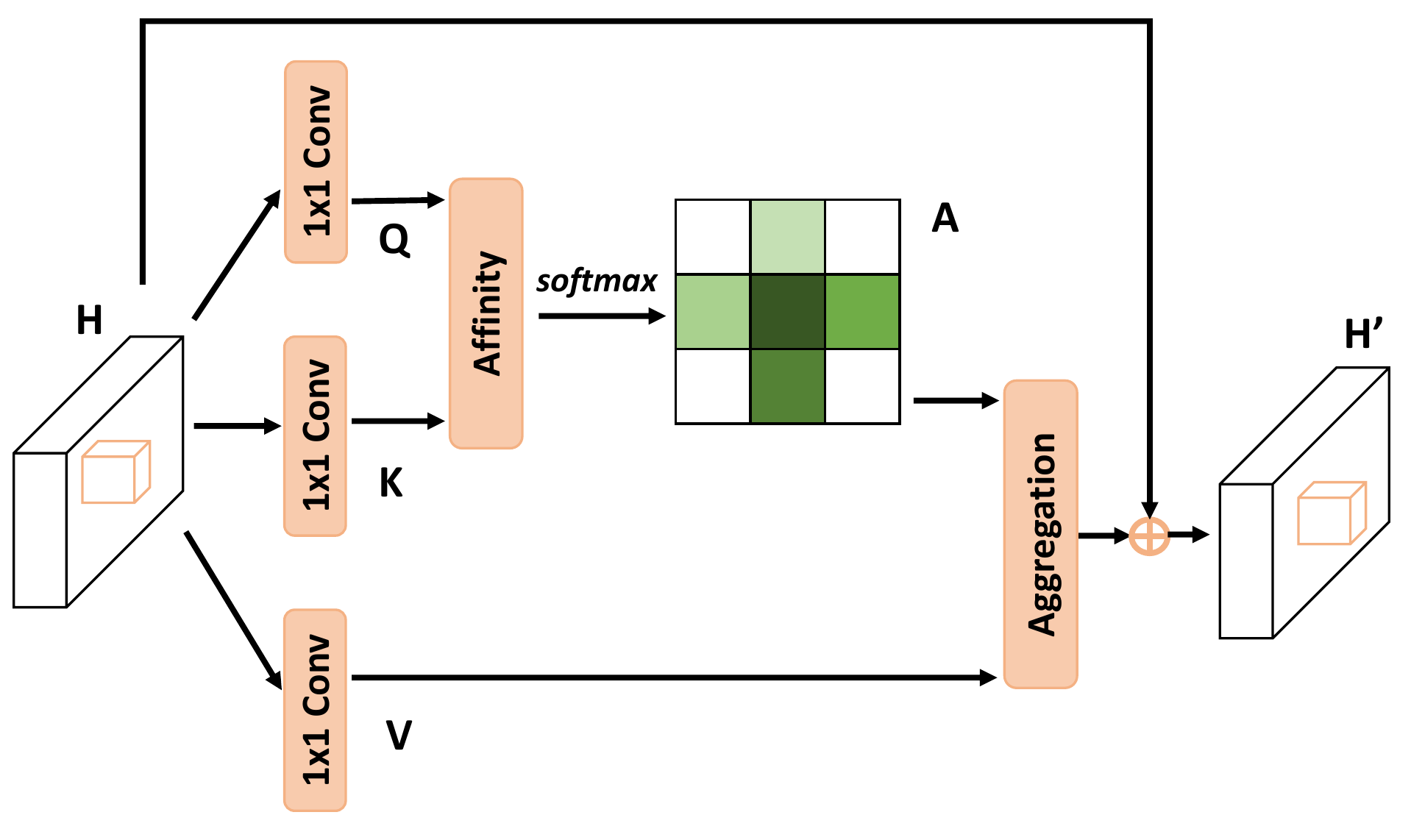}
    \caption{The details of criss-cross attention module.}
    \label{fig:cca_module}
\end{figure}

To model full-image dependencies over local feature representations using light-weight computation and memory, we introduce a criss-cross attention module. The criss-cross attention module collects contextual information in horizontal and vertical directions to enhance pixel-wise representative capability. As shown in Fig.~\ref{fig:cca_module}, given a local feature map $\mathbf{H} \in \mathbb{R}^{C \times W \times H}$, the module first applies two convolutional layers with $1\times 1$ filters on $\mathbf{H}$ to generate two feature maps $\mathbf{Q}$ and $\mathbf{K}$, respectively, 
where $\{ \mathbf{Q}, \mathbf{K} \} \in \mathbb{R}^{C' \times W \times H}$. $C'$ is the number of channel, which is less than $C$ for dimension reduction.

After obtaining $\mathbf{Q}$ and $\mathbf{K}$, we further generate an attention map $\mathbf{A} \in \mathbb{R}^{(H+W-1) \times (W \times H)}$ via \textbf{Affinity} operation. At each position $\textbf{u}$ in the spatial dimension of $\mathbf{Q}$, we can obtain a vector $\mathbf{Q_u} \in \mathbb{R}^{C'}$. Meanwhile, we can also obtain the set $\mathbf{\Omega_u} \in \mathbb{R}^{(H+W-1) \times C'}$ by extracting feature vectors from $\mathbf{K}$ which are in the same row or column with position $\textbf{u}$. $\mathbf{\Omega}_{i,\textbf{u}} \in \mathbb{R}^{C'}$ is the $i$-th element of $\mathbf{\Omega_u}$.
The \textbf{Affinity} operation is then defined as follows.
\begin{align} \label{eq:Affinity}
d_{i,\textbf{u}} = \mathbf{Q_u}\mathbf{\Omega}_{i,\textbf{u}}^\intercal,
\end{align}
where $d_{i,\textbf{u}} \in \mathbf{D}$ is the degree of correlation between features $\mathbf{Q_u}$ and $\mathbf{\Omega}_{i,\textbf{u}}$, $i=[1,...,H+W-1]$, and $\mathbf{D} \in \mathbb{R}^{(H+W-1) \times (W \times H)}$. Then, we apply a softmax layer on $\mathbf{D}$ over the channel dimension to calculate the attention map $\mathbf{A}$.

Another convolutional layer with $1\times 1$ filters is applied on $\mathbf{H}$ to generate $\mathbf{V} \in \mathbb{R}^{C \times W \times H}$ for feature adaptation. At each position $\textbf{u}$ in the spatial dimension of $\mathbf{V}$, we can obtain a vector $\mathbf{V_u} \in \mathbb{R}^{C}$ and a set $\mathbf{\Phi_u} \in \mathbb{R}^{(H+W-1) \times C}$. The set $\mathbf{\Phi_u}$ is a collection of feature vectors in $\mathbf{V}$  which are in the same row or column with position $u$. The contextual information is collected by an \textbf{Aggregation} operation defined as follows.
\begin{align} \label{eq:Aggregation}
    \mathbf{H_u'} = 
    \sum\limits_{i=0}^{H+W-1} \mathbf{A}_{i,\textbf{u}}\mathbf{\Phi_{i,u}} + \mathbf{H_u},
\end{align}
where $\mathbf{H_u'}$ is a feature vector in $\mathbf{H'} \in \mathbb{R}^{C \times W \times H}$ at position $u$ and $\mathbf{A}_{i,\textbf{u}}$ is a scalar value at channel $i$ and position $\textbf{u}$ in $\mathbf{A}$. The contextual information is added to local feature $\mathbf{H}$ to augment the pixel-wise representation. Therefore, it has a wide contextual view and selectively aggregates contexts according to the spatial attention map. These feature representations achieve mutual gains and are
more robust for semantic segmentation.

\subsection{Recurrent Criss-Cross Attention (RCCA)}

Despite the criss-cross attention module can capture contextual information in horizontal and vertical directions, the connections between one pixel and its around ones that are not in the criss-cross path are still absent. To tackle this problem, we innovatively and simply introduce a RCCA operation based on the criss-cross attention. The RCCA module can be unrolled into $R$ loops. In the first loop, the criss-cross attention takes the feature map $\mathbf{H}$ extracted from a CNN model as the input and output the feature map $\mathbf{H'}$, where $\mathbf{H}$ and $\mathbf{H'}$ are with the same shape. In the second loop, the criss-cross attention takes the feature map $\mathbf{H'}$ as the input and output the feature map $\mathbf{H''}$. As shown in Fig.~\ref{fig:architecture}, the RCCA module is equipped with two loops ($R=2$) which is able to harvest full-image contextual information from all pixels to generate new features with dense and rich contextual information. 

\begin{figure}[!t]
    \centering
    \includegraphics[width=1.0\linewidth]{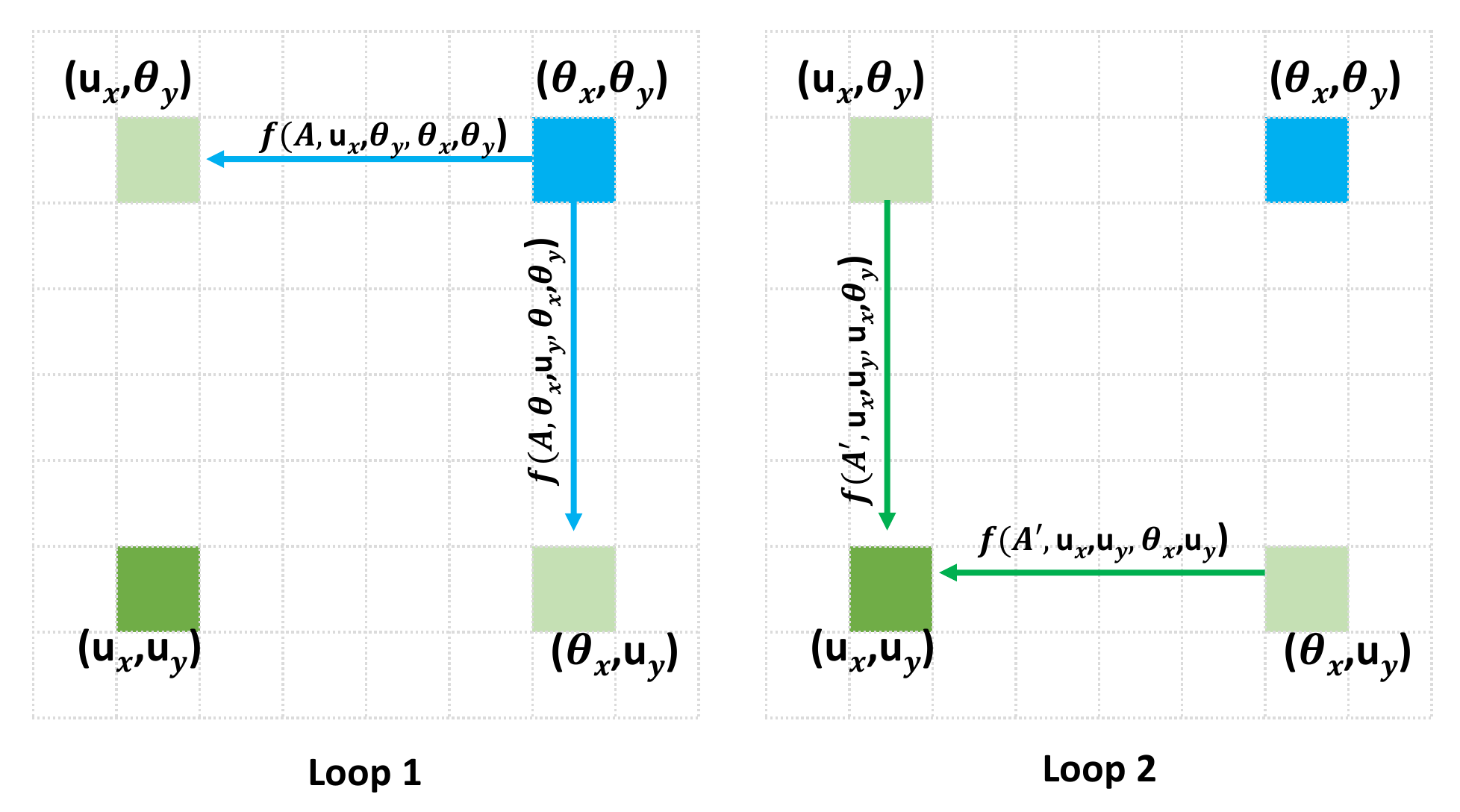}
    \caption{An example of information propagation when the loop number is 2.}
    \label{fig:alg2}
\end{figure}

We denote $\mathbf{A}$ and $\mathbf{A'}$ as the attention maps in loop 1 and loop 2, respectively. Since we are interested only in contextual information spreads in spatial dimension rather than in channel dimension, the convolutional layer with $1\times 1$ filters can be view as the identical connection. 
In the case of $R = 2$, the connections between any two spatial positions in the feature map built up by the RCCA module can be clearly and quantitatively described by introducing function $f$ defined as follows.
\begin{equation*}
    \exists i \in \mathbb{R}^{H + W - 1}, s.t. \ \mathbf{A}_{i, \mathbf{u}} = f(\mathbf{A}, u^{CC}_x, u^{CC}_y, u_x, u_y), 
\end{equation*}{}
where $\mathbf{u}(u_x, u_y) \in \mathbb{R}^{H \times W}$ is any spatial position in
$\mathbf{H}$ and $\mathbf{u}^{CC}(u^{CC}_x, u^{CC}_y) \in \mathbb{R}^{H + W - 1}$ is a position in the criss-cross structure centered at $\mathbf{u}$. The function $f$ is actually an \textbf{one-to-one mapping} from the position pair $(\mathbf{u}^{CC}, \mathbf{u}) \in \mathbb{R}^{(H + W - 1) \times (H \times W)}$ in the feature map to a particular element $\mathbf{A}_{i, \mathbf{u}} \in \mathbb{R}^{(H + W - 1) \times (H \times W)}$ in the attention map $\mathbf{A} \subset \mathbb{R}^{(H + W - 1) \times (H \times W)}$, where $\mathbf{u}^{CC}$ maps to a particular row $i$ in  $\mathbf{A}$ and $\mathbf{u}$ maps to a particular column in $\mathbf{A}$.

\red{With the help of function $f$, we can easily describe the information propagation between any position $\mathbf{u}$ in $\mathbf{H}^{\prime \prime}$ and any position $\boldsymbol{\theta}$ in $\mathbf{H}$. It is obvious that information could flow from $\boldsymbol{\theta}$ to $\mathbf{u}$ when $\boldsymbol{\theta}$ is in the criss-cross path of $\mathbf{u}$. }

\red{Then, we focus on another situation in which $\boldsymbol{\theta}(\theta_x, \theta_y)$ is NOT in the criss-cross path of $\mathbf{u}(u_x, u_y)$. To make it easier to understand, we visualize the information propagation in Fig.~\ref{fig:alg2}. The position $(\theta_x, \theta_y)$, which is blue, firstly passes the information into the $(u_x, \theta_y)$ and $(\theta_x, u_y)$ (light green) in the loop 1. The propagation could be quantified by function $f$. It should be noted that 
these two points $(u_x, \theta_y)$ and $(\theta_x, u_y)$ are in the criss-cross path of $\mathbf{u}(u_x, u_y)$. Then, the positions $(u_x, \theta_y)$ and $(\theta_x, u_y)$ pass the information into the $(u_x, u_y)$ (dark green) in the loop 2. Thus, the information in $\boldsymbol{\theta}(\theta_x, \theta_y)$ could eventually flow into $\mathbf{u}(u_x, u_y)$ even if $\boldsymbol{\theta}(\theta_x, \theta_y)$ is NOT in the criss-cross path of $\mathbf{u}(u_x, u_y)$.}

In general, our RCCA module makes up for the deficiency of criss-cross attention that cannot obtain the dense contextual information from all pixels. Compared with criss-cross attention, the RCCA module ($R=2$) does not bring extra parameters and can achieve better performance with the cost of a minor computation increment. 

\subsection{Learning Category Consistent Features} 
For semantic segmentation tasks, the pixels belonging to the same category should have the similar features, while the pixels from different categories should have far apart features. We name such a characteristic as category consistency. The deep features produced by RCCA have full-image context; however, the aggregated feature may have the problem of over-smoothing, which is a common issue in graph neural networks. To address this potential issue, beside the cross-entropy loss $\ell_{seg}$ to penalize the mismatch between the final predicted segmentation maps and ground truth, we further introduce the category consistent loss to drive RCCA module to learn category consistent features directly.

In \cite{de2017semantic}, a discriminative loss function with three competing terms is proposed for instance segmentation. In particular, the three terms, denoted as $\ell_{var}, \ell_{dis}, \ell_{reg}$, are adopted to 1) penalize large distances between features with the same label for each instance, 2) penalize small distances between the mean features of different labels, and 3) draw mean features of all categories towards the origin, respectively.

Motivated by \cite{de2017semantic}, we first adapt a discriminative loss for semantic segmentation rather than instance segmentation, then replace the first term with more robust one: instead of using quadratic function as the distance function to penalize mismatch all along, we design a piece-wise distance function to make the optimization more robust.

Let $C$ be the set of classes that are present in the mini-batch images. $N_c$ is the number of valid elements belonging to category $c \in C$. $h_i \in \textbf{H}$ is the feature vector at spatial position $i$. $\mu_c$ is the mean feature of category $c \in C$ (the cluster center). $\varphi$ is a piece-wise distance function. $\delta_v$ and $\delta_d$ are respectively the margins. \red{
In particular, Eq.~\ref{eq:phi_var} is a piece-wise distance function and the function $\varphi_{var}$ will be zero, quadratic, and linear function when the distance from the center $\mu_c$ is within $d_v$, in range of $(\delta_v, \delta_d]$, and exceeds $\delta_d$, respectively.}

    \begin{align} \label{eq:variance}
    \ell_{var} = \frac{1}{|C|}\sum_{c \in C}\frac{1}{N_c}\sum_{i=1}^{N_c}\varphi_{var}(h_i, \mu_c),
    \end{align}
    
    \begin{align} \label{eq:distance}
    \ell_{dis} = \frac{1}{|C|(|C|-1)}\underset{c_a \neq c_b}{\sum_{c_a \in C}\sum_{c_b \in C}}\varphi_{dis}(\mu_{c_a}, \mu_{c_b}),
    \end{align}
    
    \begin{align} \label{eq:variance}
    \ell_{reg} = \frac{1}{|C|}\sum_{c \in C}\|\mu_c\|,
    \end{align}
    
    \begin{equation} \label{eq:phi_var}
    \varphi_{var} =\left\{
    \begin{array}{l}
    \|\mu_c-h_i\| - \delta_d + (\delta_d-\delta_v)^2, \quad {\|\mu_c-h_i\|> \delta_d}\\
    (\|\mu_c-h_i\| - \delta_v)^2, \qquad\quad {\delta_v<\|\mu_c-h_i\|\leq \delta_d}\\
    0, \qquad\qquad\qquad\qquad\qquad\qquad\quad { \|\mu_c-h_i\| \leq \delta_v}\\
    \end{array} \right.
    \end{equation}
    
    \begin{equation} \label{eq:phi_dis}
    \varphi_{dis} =\left\{
    \begin{array}{ll}
    (2\delta_d-\|\mu_{c_a}-\mu_{c_b}\|)^2, & {\|\mu_{c_a}-\mu_{c_b}\| \leq 2\delta_d}\\
    0, & { \|\mu_{c_a}-\mu_{c_b}\| > 2\delta_d}\\
    \end{array} \right.
    \end{equation}

To reduce the computation load, we first apply a convolutional layer with $1\times 1$ filters on the output of RCCA module for dimension reduction and then apply these three loss on the feature map with fewer channels. The final loss $\ell$ is weighted sum of all losses.
\begin{align} \label{eq:variance}
    \mathbf{\ell} = \ell_{seg}+\alpha\ell_{var}+\beta\ell_{dis}+\gamma\ell_{reg},
\end{align}
where $\alpha$, $\beta$ and $\ell$ are the weight parameters. In our experiments we set $\delta_v=0.5$, $\delta_d=1.5$, $\alpha=\beta=1$, $\gamma=0.001$ and $16$ as the number of channels for dimension reduction. 

\subsection{3D Criss-Cross Attention}

\begin{figure}[!tp]
    \centering
    \includegraphics[width=1.0\linewidth]{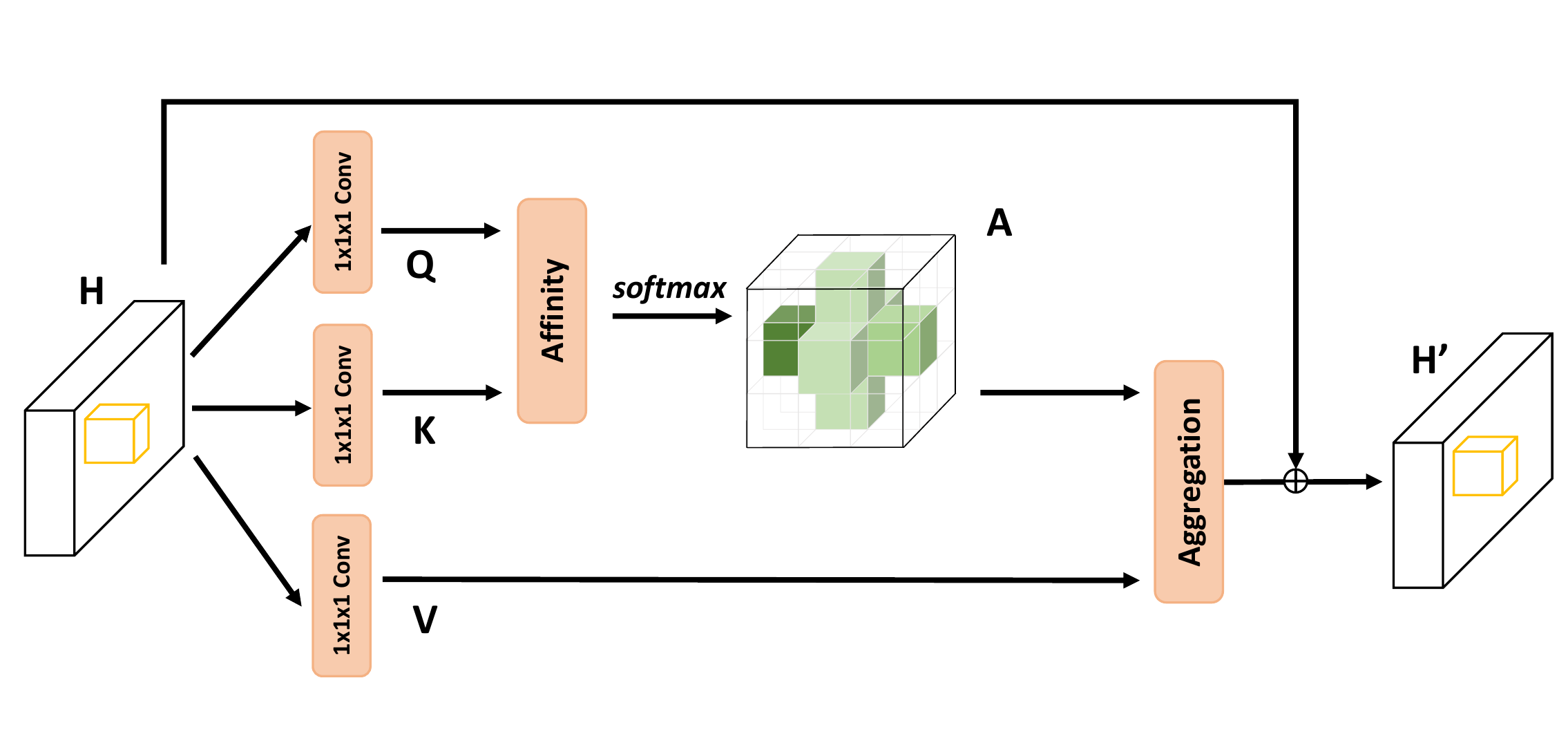}
    \caption{The details of 3D criss-cross attention module.}
    \label{fig:cca_module3d}
\end{figure}

To adapt our method from 2D applications to 3D dense prediction tasks, we introduce 3D Criss-Cross Attention. In general, the architecture of 3D Criss-Cross Attention is an extension the 2D version by additional collecting more contextual information from the temporal dimension. As shown in Fig.~\ref{fig:cca_module3d}, given a local feature map $\mathbf{H} \in \mathbb{R}^{C \times T \times W \times H}$, where $T$ is axial dimension (\ie, temporal dimension in video data). The module firstly applies two convolutional layers with $1\times 1 \times 1$ filters on $\mathbf{H}$ to generate two feature maps $\mathbf{Q}$ and $\mathbf{K}$, respectively, where $\{ \mathbf{Q}, \mathbf{K} \} \in \mathbb{R}^{C' \times T \times W \times H}$. 

After obtaining the feature maps $\mathbf{Q}$ and $\mathbf{K}$, we further generate an attention map $\mathbf{A} \in \mathbb{R}^{(T+H+W-2) \times T \times W \times H}$ via the \textbf{Affinity} operation. At each position $u$ of $\mathbf{Q}$, we can obtain a vector $\mathbf{Q_u} \in \mathbb{R}^{C'}$. $u$ contains three coordinate values $(t,x,y)$.
We can also obtain the set $\mathbf{\Omega_u} \in \mathbb{R}^{(T+H+W-2) \times C'}$ by extracting feature vectors from $\mathbf{K}$ with at least two coordinate values equal to $u$. 
$\mathbf{\Omega_{i,u}} \in \mathbb{R}^{C'}$ is the $i$-th element of $\mathbf{\Omega_u}$.
The \textbf{Affinity} operation is then defined as follows.
    \begin{align} \label{eq:Affinity}
    d_{i,u} = \mathbf{Q_u}\mathbf{\Omega_{i,u}}^\intercal,
    \end{align}
where $d_{i,u} \in \mathbf{D}$ is the degree of correlation between feature $\mathbf{Q_u}$ and $\mathbf{\Omega_{i,u}}$, $i=[1,...,(T+H+W-2)]$, $\mathbf{D} \in \mathbb{R}^{(T+H+W-2) \times T \times W \times H}$. Then, we apply a softmax layer on $\mathbf{D}$ over the first dimension to calculate the attention map $\mathbf{A}$.

Another convolutional layer with $1\times 1 \times 1$ filters is applied on $\mathbf{H}$ to generate $\mathbf{V} \in \mathbb{R}^{C \times T \times W \times H}$ for feature adaptation.
At each position $u$ in the spatial dimension of $\mathbf{V}$, we can obtain a vector $\mathbf{V_u} \in \mathbb{R}^{C}$ and a set $\mathbf{\Phi_u} \in \mathbb{R}^{(T+H+W-2) \times C}$. The the set $\mathbf{\Phi_u}$ is a collection of feature vectors in $\mathbf{V}$  which are in the criss-cross structure centered at $u$.
The contextual information is collected by the \textbf{Aggregation} operation:
    \begin{align} \label{eq:Aggregation}
    \mathbf{H_u'} = 
    \sum\limits_{i=0}^{T+H+W-2} \mathbf{A_{i,u}}\mathbf{\Phi_{i,u}} + \mathbf{H_u},
    \end{align}
where $\mathbf{H_u'}$ is a feature vector in the output feature map $\mathbf{H'} \in \mathbb{R}^{C \times T \times W \times H}$ at position $u$. $\mathbf{A_{i,u}}$ is a scalar value at channel $i$ and position $u$ in $\mathbf{A}$.

\section{Experiments} \label{Experiments}
To evaluate the effectiveness of the CCNet, we carry out comprehensive experiments on the Cityscapes dataset~\cite{cordts2016cityscapes}, the ADE20K dataset~\cite{zhou2017scene}, the COCO dataset~\cite{lin2014microsoft}, the LIP dataset~\cite{liang2018look} and the CamVid dataset~\cite{brostow2008segmentation}.  Experimental results demonstrate that CCNet achieves state-of-the-art performance on Cityscapes, ADE20K and LIP. Meanwhile, CCNet can bring constant performance gain on COCO for instance segmentation. 
In the following subsections, we first introduce the datasets and implementation details, then we perform a series of ablation experiments on Cityscapes dataset. Finally, we report our results on ADE20K, LIP, COCO and CamVid datasets.

\subsection{Datasets and Evaluation Metrics}

We adopt Mean IoU (mIOU, mean of class-wise intersection over union) for Cityscapes, ADE20K, LIP and CamVid and the standard COCO metrics Average Precision (AP) for COCO.
\begin{itemize}
    \item \textbf{Cityscapes} is tasked for urban segmentation.
    Only the 5,000 finely annotated images are used in our experiments and are divided into 2,975/500/1,525 images for training, validation, and testing, respectively.
    \item \textbf{ADE20K} is a recent scene parsing benchmark containing dense labels of 150 stuff/object categories. The dataset includes 20k/2k/3k images for training, validation and testing, respectively.
    \item \textbf{LIP} is a large-scale single human parsing dataset. There are 50,462 images with fine-grained annotations at pixel-level with 19 semantic human part labels and one background label. Those images are further divided into 30k/10k/10k for training, validation and testing, respectively.
    \item \textbf{COCO} is a very challenging dataset for instance segmentation that contains 115k images over 80 categories for training, 5k images for validation and 20k images for testing.
    \item \textbf{CamVid} is one of the datasets focusing on semantic segmentation for autonomous driving scenarios. It is composed of
    701 densely annotated images with size $720 \times 960$ from five video sequences.
\end{itemize}

\subsection{Implementation Details}

\noindent\textbf{Network Structure} 
For semantic segmentation, we choose the ImageNet pre-trained ResNet-101~\cite{he2016deep} as our backbone network, remove its last two down-sampling operations, and employ dilated convolutions in the subsequent convolutional layers following the previous work~\cite{chen2014semantic}, resulting in the output stride as 8. For human parsing, we choose CE2P~\cite{ruan2019devil} as our baseline and replace the  Context Embedding module with RCCA. For instance segmentation, we choose Mask-RCNN~\cite{he2017mask} as our baseline. For video semantic segmentation, we also choose Cityscapes pre-trained ResNet-101~\cite{he2016deep} as our backbone network with 3D RCCA.

\vspace{1em}
\noindent\textbf{Training settings}
SGD with mini-batch is used for training. For semantic segmentation, the initial learning rate is 1e-2 for Cityscapes and ADE20K. Following the prior works~\cite{chen2018deeplab, zhang2018context}, we employ a poly learning rate policy where the initial learning rate is multiplied by $ 1 - (\frac{iter}{max\_iter})^{power} $ with $power$ = 0.9. We use the momentum of 0.9 and a weight decay of 0.0001. For Cityscapes, the training images are augmented by randomly scaling (from 0.75 to 2.0), then randomly cropping out high-resolution patches ($769 \times 769$) from the resulting images. Since the images from ADE20K are with various sizes, we adopt an augmentation strategy of resizing the short side of input image to a length randomly chosen from the set \{300, 375, 450, 525, 600\}. For human parsing, the model are trained and tested with the input size of $473\times473$. For instance segmentation, we take the same training settings as that of Mask-RCNN~\cite{he2017mask}. For video semantic segmentation, we sample 5 temporally ordered frames from a training video as training data and the input size is $504\times504$.

\subsection{Experiments on Cityscapes}

\subsubsection{Comparisons with state-of-the-arts}
Results of other state-of-the-art semantic segmentation solutions on Cityscapes are summarized in Tab.~\ref{tab:cityscape_val_test}. For val set, we provide these results for reference and emphasize that these results should not be simply compared with our method, since these methods are trained on different (even larger) training sets or different basic network. Among these approaches, Deeplabv3~\cite{chen2017rethinking} adopts multi-scale testing strategy. Deeplabv3+~\cite{chen2018encoder} and DPC~\cite{chen2018searching} both use a more stronger backbone (\ie, Xception-65 \& 71 \emph{vs.} ResNet-101). In addition, DPC~\cite{chen2018searching} makes use of additional dataset, \ie, COCO, for pre-training beyond the training set of Cityscapes. The results show that the proposed CCNet with single-scale testing still achieve comparable performance without bells and whistles. 

Additionally, we also train the best learned CCNet with ResNet-101 as the backbone using both training and validation sets and make the evaluation on the test set by submitting our test results to the official evaluation server. Most of methods~\cite{chen2018deeplab, lin2017refinenet, zhang2017scale, peng2017large, wang2018understanding, zhao2017pyramid, yu2018bisenet, ke2018adaptive, zhao2018psanet, yu2018learning} adopt the same backbone as ours and the others~\cite{wu2016wider,yang2018denseaspp} utilize stronger backbones. 
From Tab.~\ref{tab:cityscape_val_test}, it can be observed that our CCNet substantially outperforms all the previous state-of-the-arts on test set.
Among the approaches, PSANet~\cite{zhao2018psanet} is the most related to our method which generates sub attention map for each pixel. One of the differences is that the sub attention map has $2 \times H \times W$ weights in PSANet and  $H + W - 1$ weights in CCNet. Even with lower computation cost and memory usage, our method still achieves better performance.
    
    \begin{table}[!t]
        \renewcommand{\arraystretch}{1.3}
        \setlength{\tabcolsep}{0.9em}
        \caption{Comparison with state-of-the-arts on 
        Cityscapes (test).}
        \label{tab:cityscape_val_test}
        \centering \small
        \begin{threeparttable}
        \begin{tabular}{|l|c|c|}
            \hline
            Method & Backbone & mIOU(\%)  \\
            \hline
            \multicolumn{3}{|l|}{\textit{Performance on val set}} \\
            \hline
            DeepLabv3~\cite{chen2017rethinking} & ResNet-101 & 79.3  \\
            DeepLabv3+~\cite{chen2018encoder} & Xception-65 & 79.1\\
            DPC~\cite{chen2018searching}~\dag & Xception-71 & 80.8\\
            CCNet & ResNet-101 & 80.5\\
            \hline
            \multicolumn{3}{|l|}{\textit{Performance on test set}} \\
            \hline
            DeepLab-v2~\cite{chen2018deeplab} & ResNet-101 & 70.4  \\
            RefineNet~\cite{lin2017refinenet}~\ddag & ResNet-101 & 73.6\\
            SAC~\cite{zhang2017scale}~\ddag & ResNet-101 & 78.1\\
            GCN~\cite{peng2017large}~\ddag  & ResNet-101 & 76.9\\
            DUC~\cite{wang2018understanding}~\ddag & ResNet-101 & 77.6\\
            ResNet-38~\cite{wu2016wider} & WiderResnet-38 & 78.4\\
            PSPNet~\cite{zhao2017pyramid} & ResNet-101 & 78.4 \\
            BiSeNet~\cite{yu2018bisenet}~\ddag & ResNet-101 & 78.9 \\
            AAF~\cite{ke2018adaptive} & ResNet-101 & 79.1 \\
            PSANet~\cite{zhao2018psanet}~\ddag & ResNet-101 & 80.1 \\
            DFN~\cite{yu2018learning}~\ddag & ResNet-101 & 79.3 \\
            DenseASPP~\cite{yang2018denseaspp}~\ddag & DenseNet-161 & 80.6\\
            CCNet~\ddag & ResNet-101 & \textbf{81.9}\\
            \hline
        \end{tabular}
        \begin{tablenotes} 
        \item \dag ~use extra COCO dataset for training.
        \item \ddag ~train with both the train-fine and val-fine datasets.
      \end{tablenotes}
      \end{threeparttable}
      \vspace{-0mm}
    \end{table}

\begin{figure*}[!t]
    \centering
    \includegraphics[width=1.0\linewidth]{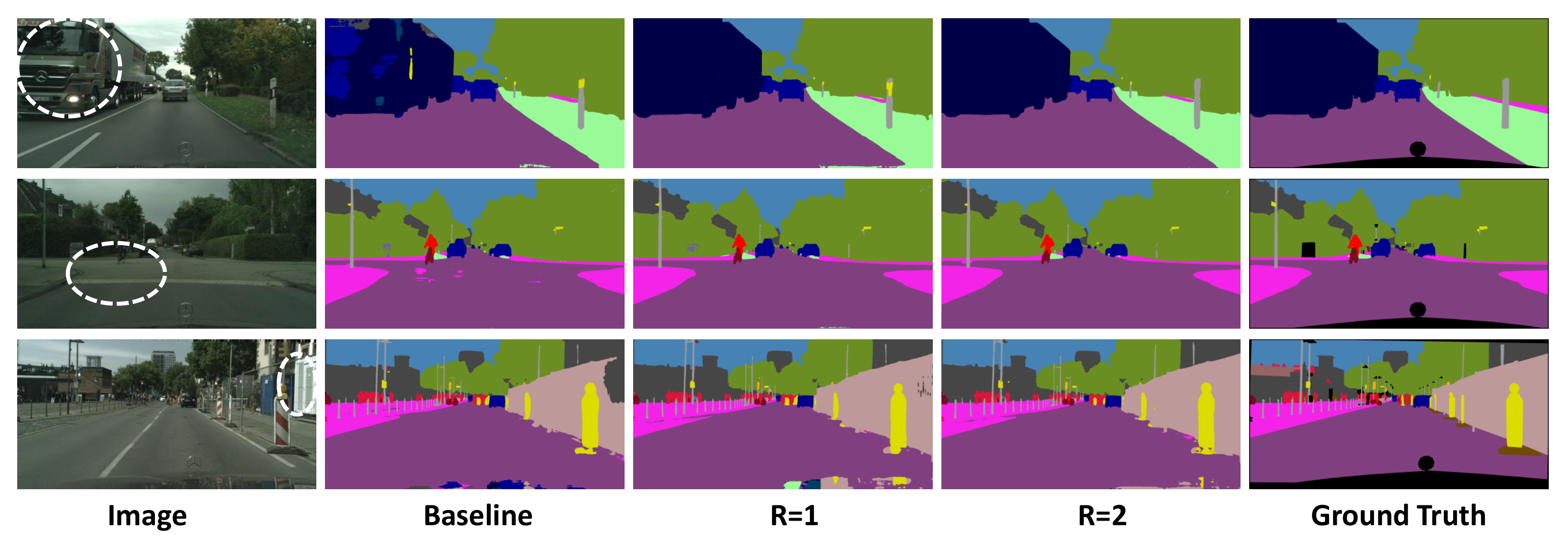}
    \caption{Visualization results of RCCA with different loops on Cityscapes validation set.}
    \label{fig:val_r}
    \vspace{-0mm}
\end{figure*}

\subsubsection{Ablation studies}
To verify the rationality of the CCNet, we conduct extensive ablation experiments on the validation set of Cityscapes with different settings for CCNet.

\begin{table}[!t]
    \renewcommand{\arraystretch}{1.3}
    \setlength{\tabcolsep}{0.6em}
    \caption{Performance on Cityscapes (val) for different number of loops in RCCA. FLOPs and memory increment are estimated for an input of $1 \times 3 \times 769 \times 769$.}
    \label{tab:ablation_r}
    \centering \small
    \begin{tabular}{|l|c|c|c|}
        \hline
        Loops & GFLOPs($\blacktriangle$) & Memory(M$\blacktriangle$) & mIOU(\%)  \\
        \hline
        baseline & 0 & 0 & 75.1 \\
        $R=1$ & 8.3 & 53 & 78.0 \\
        $R=2$ & 16.5 & 127 & 79.8 \\
        $R=3$ & 24.7 & 208 & 80.2 \\
        \hline
    \end{tabular}
\end{table}
    
\vspace{1em}
\noindent\textbf{The effect of the RCCA module}
Tab.~\ref{tab:ablation_r} shows the performance on the Cityscapes validation set by adopting different number of loop in RCCA. All experiments are conducted using ResNet-101 as the backbone. Besides, the input size of training images is $769 \times 769$ and the size of the input feature map $\textbf{H}$ of RCCA is $97 \times 97$. Our baseline network is the ResNet-based FCN with dilated convolutional module incorporated at stage 4 and 5, \ie, dilation rates are set to 2 and 4 for these two stages respectively. The increment of FLOPs and memory usage are estimated when $R=1,2,3$, respectively. 

We observe that adding a criss-cross attention module into the baseline, donated as $R=1$, improves the performance by 2.9\%, which can effectively demonstrates the significance of criss-cross attention. Furthermore, increasing the number of loops from 1 to 2 can further improve the performance by 1.8\%, demonstrating the effectiveness of dense contextual information. Finally, increasing loops from 2 to 3 slightly improves the performance by 0.4\%. Meanwhile, with the increasing the number of loops, the FLOPs and  usage of GPU memory keep increasing. These results prove that the proposed criss-cross attention can significantly improve the performance by capturing contextual information in horizontal and vertical direction. In addition, the proposed RCCA is effective in capturing the dense and global contextual information, which can finally benefit the performance of semantic segmentation.
To balance the performance and resource usage, we choose $R=2$ as default settings in all the following experiments. 

To further validate the effectiveness of the criss-cross module, we provide the qualitative comparisons in Fig.~\ref{fig:val_r}. We leverage the \emph{white} \emph{circles} to indicate those challenging regions that are easily to be misclassified. It can be seen that these challenging regions are progressively corrected with the increasing the number of loops, which can well prove the effectiveness of dense contextual information aggregation for semantic segmentation. \\

\begin{table}[!t]
    \renewcommand{\arraystretch}{1.3}
    \setlength{\tabcolsep}{0.6em}
    \caption{Performance on Cityscapes (val) for different kinds of category consistent loss.}
    \label{tab:ablation_ccl_type}
    \centering \small
    \begin{tabular}{|l|c|c|}
        \hline
        Function Type & Successes & Mean mIOU(\%)  \\
        \hline
        Quadratic function & 6/10 &  79.2 \\
        Piece-wise function & 9/10 & 79.3 \\
        \hline
    \end{tabular}
\end{table}

\noindent\textbf{The effect of the category consistent loss}
Tab.~\ref{tab:ablation_comparison} also shows the performance on the Cityscapes validation set by adopting the proposed category consistent loss. The category consistent loss is donated as ``CCL'' in the table. As we can see, adopting the category consistent loss could stably bring \~0.7\% mIoU gain with both Resnet-101 and Resnet-50, which prove the effectiveness of the proposed category consistent loss for semantic segmentation. \red{To prove that the proposed piece-wise function is more robust than the original one, we conduct 10 times of the training processes using ResNet-50 for each kind of loss function. The training is deemed to fail when the loss value is NaN, thus we can calculate the success rate (number of successful training / total number of training). The experimental results in Table~\ref{tab:ablation_ccl_type} demonstrate that using the piece-wise function has higher training success rate than using the original one. Besides, using the piece-wise function could achieve slightly better performance than a single quadratic function. Because we relax the punishment in the Eq.~\ref{eq:phi_var} to reduce the numerical values and gradients especially when the distance from the center exceeds $\delta_d$. This relaxation makes the optimization much more stable. }

    \begin{table}[!t]
        \renewcommand{\arraystretch}{1.3}
        \setlength{\tabcolsep}{1.0em}
        \caption{Comparison of context aggregation approaches on Cityscapes (val).}
        \label{tab:ablation_comparison}
        \centering \small
        \begin{tabular}{|l|c|}
            \hline
            Method & mIOU(\%)  \\
            \hline
            ResNet101-Baseline & 75.1 \\
            ResNet101+GCN & 78.1 \\
            ResNet101+PSP & 78.5 \\
            ResNet101+ASPP & 78.9 \\
            ResNet101+NL & 79.1 \\
            ResNet101+RCCA(R=2)  & 79.8\\
            ResNet101+RCCA(R=2)+CCL  & \bf{80.5}\\
            \hline \hline
            ResNet50-Baseline & 73.3 \\
            ResNet50+GCN & 76.2 \\
            ResNet50+PSP & 76.4 \\
            ResNet50+ASPP & 77.1 \\
            ResNet50+NL & 77.3 \\
            ResNet50+HV & 77.3 \\
            ResNet50+HV\&VH & 77.8 \\
            ResNet50+RCCA(R=2)  & 78.5\\
            ResNet50+RCCA(R=2)+CCL  & \bf{79.3}\\
            \hline
        \end{tabular}
    \end{table}
    
    \begin{figure*}[!t]
        \centering
        \includegraphics[width=1.0\linewidth]{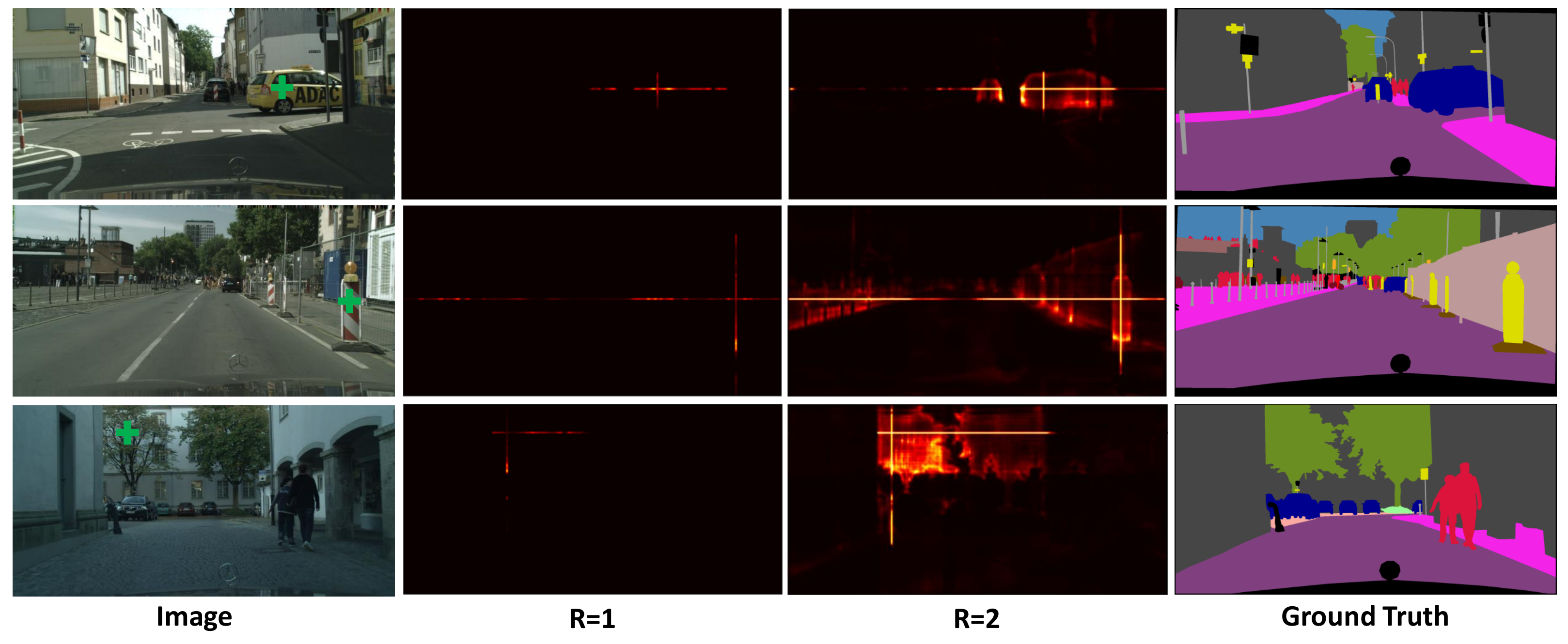}
        \caption{Visualization of attention module on Cityscapes validation set. The left column is the input images, the 2 and 3 columns are pixel-wise attention maps when $R=1$ and $R=2$ in RCCA. }
        \label{fig:attention_vis}
    \end{figure*}

\vspace{1em}
\noindent\textbf{Comparison of other context aggregation approaches}
We compare the performance of several different context aggregation approaches on the Cityscapes validation set with ResNet-50 and ResNet-101 as backbone networks.

Specifically, the baselines of context aggregation mainly include: 1) Peng \etal~\cite{peng2017large} utilized global convolution filters for contextual information aggregation, donated as ``+GCN''. 2) Zhao \etal ~\cite{zhao2017pyramid} proposed Pyramid pooling which is the simple and effective way to capture global contextual information, donated as ``+PSP''; 3) Chen \etal~\cite{chen2017rethinking} used different dilation convolutions to harvest pixel-wise contextual information at the different range, donated as ``+ASPP''; 4) Wang \etal~\cite{wang2018non} introduced non-local network for context aggregation, donated as ``+NL''.

In Tab.~\ref{tab:ablation_comparison}, both ``+NL" and ``+RCCA" achieve better performance compared with the other context aggregation approaches, which demonstrates the importance of capturing full-image contextual information.
More interestingly, our method achieves better performance than ``+NL''. This reason may be attributed to the sequentially recurrent operation of criss-cross attention. Concretely, ``+NL'' generates an attention map directly from the feature which has limit receptive field and short-range dependencies. In contrast, our ``+RCCA'' takes two steps to form dense contextual information, leading to that the latter step can learn a better attention map benefiting from the feature map produced by the first step in which some long-range dependencies has already been embedded.

To prove the effectiveness of attention with criss-cross shape, we compare criss-cross shape with other shapes in Tab.~\ref{tab:ablation_comparison}. ``+HV'' means stacking horizontal attention and vertical attention. ``+HV\&VH'' means summing up features of two parallel branches, i.e. ``HV'' and ``VH''. 

We further explore the amount of computation and memory footprint of RCCA. As shown in Tab.~\ref{tab:ablation_resource_usage}, compared with ``+NL'' method,  the proposed ``+RCCA'' requires $11\times$ less GPU memory usage and significantly reduces FLOPs by about 85\% of non-local block in computing full-image dependencies, which shows that CCNet is an efficient way to capture full-image contextual information in the least amount of computation and memory footprint. \red{To further prove the effectiveness of the recurrent operation, we also run non-local module in the recurrent way, donated as ``+NL(R=2)''. As we can seen, the recurrent operation can bring more than 1 point gain. Because the recurrent operation leads to that the latter step can learn a better attention map benefiting from the feature map produced by the first step in which some long-range dependencies has already been embedded. However, compared with ``+RCCA'', ``+NL(R=2)'' needs huge GPU  memory usage, which limits the use of self-attention.}

\vspace{1em}
\noindent\textbf{Visualization of Attention Map}
To get a deeper understanding of our RCCA, we visualize the learned attention masks as shown in Fig.~\ref{fig:attention_vis}.  For each input image, we select one point (cross in green) and show its corresponding attention maps when $R=1$ and $R=2$ in columns 2 and 3, respectively. It can be observed that only contextual information from the criss-cross path of the target point is captured when $R=1$. By adopting one more criss-cross module, \ie, $R=2$, RCCA can finally aggregate denser and richer contextual information compared with that of $R=1$. Besides, we observe that the attention module could capture semantic similarity and full-image dependencies. 
    
    \begin{table}[!t]
        \renewcommand{\arraystretch}{1.3}
        \setlength{\tabcolsep}{0.4em}
        \caption{Comparison of Non-local module and RCCA. FLOPs and memory increment are estimated for an input of $1 \times 3 \times 769 \times 769$.}
        \label{tab:ablation_resource_usage}
        \centering \small
        \begin{tabular}{|l|c|c|c|}
            \hline
            Method & GFLOPs($\blacktriangle$) & Memory(M$\blacktriangle$) & mIOU(\%)  \\
            \hline
            baseline & 0 & 0 & 73.3 \\
            +NL & 108 & 1411 & 77.3 \\
            +NL(R=2) & 216 & 2820 & 78.7 \\
            +RCCA(R=2) & 16.5 & 127 & 78.5\\
            \hline
        \end{tabular}
    \end{table}
    
    \begin{figure}[!t]
        \centering
        \includegraphics[width=1.0\linewidth]{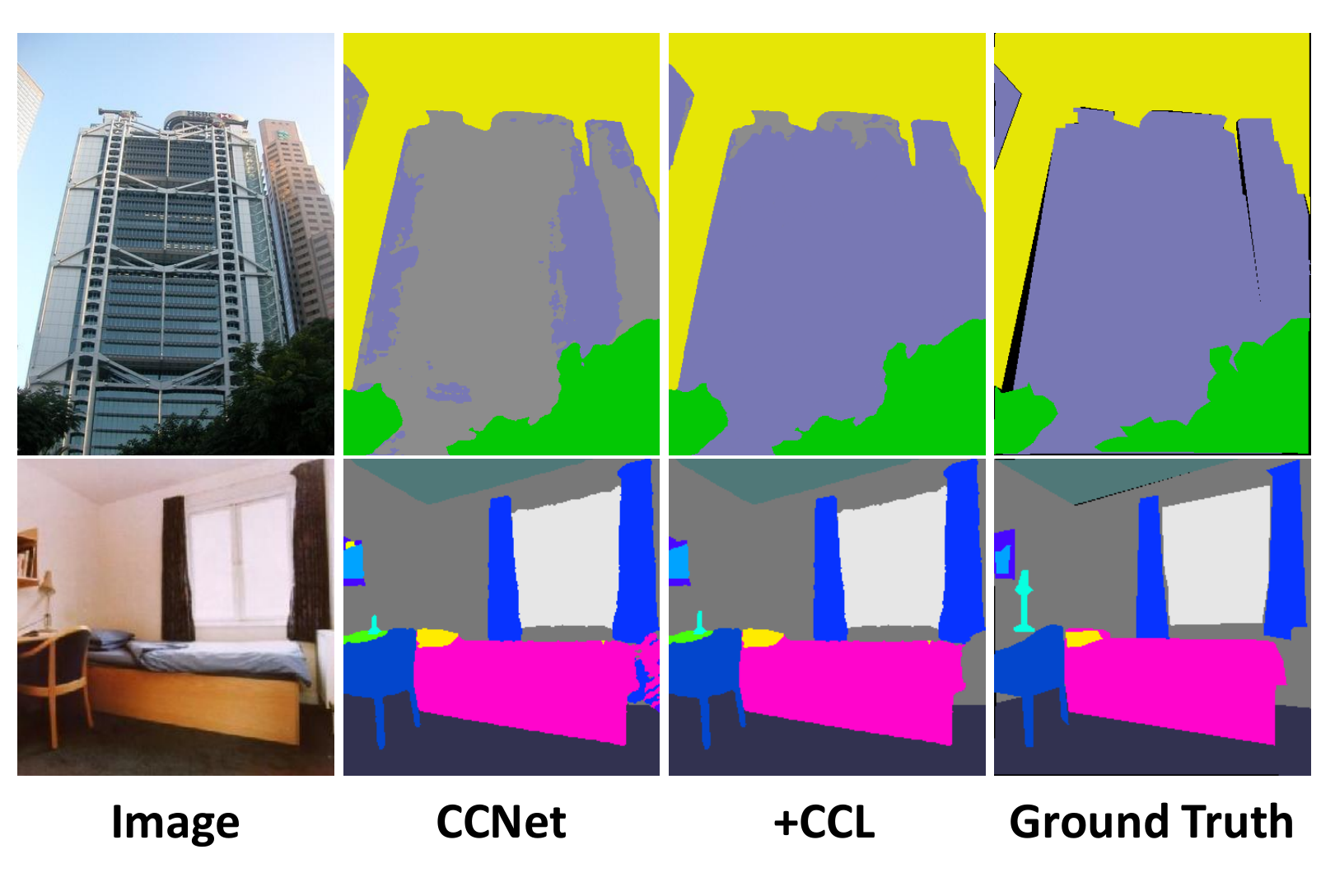}
        \caption{Visualized examples on ADE20K val set with/without category consistent loss (CCL).}
        \label{fig:ade20k_vis}
        \vspace{-5mm}
    \end{figure}

\subsection{Experiments on ADE20K}

In this subsection, we conduct experiments on the AED20K dataset, which is a very challenging scene parsing dataset. As shown in Tab.~\ref{tab:ade20k}, CCNet with CCL achieves the state-of-the-art performance of 45.76\%, outperforms the previous state-of-the-art methods by more than 1.1\% and also outperforms the conference version CCNet by 0.5\%. Some successful segmentation results are given in Fig~\ref{fig:ade20k_vis}.
Among the approaches, most of methods~\cite{zhang2017scale, zhao2017pyramid, zhao2018psanet, liang2018dynamic, xiao2018unified, zhang2018context} adopt the ResNet-101 as backbone and RefineNet~\cite{lin2017refinenet} adopts a more powerful network, \ie, ResNet-152, as the backbone. EncNet~\cite{zhang2018context} achieves previous best performance among the methods and utilizes global pooling with image-level supervision to collect image-level context information. In contrast, our CCNet adopts an alternative way to integrate contextual information by capture full-image dependencies and achieve better performance. 

    \begin{figure}[!t]
        \centering
        \includegraphics[width=1.0\linewidth]{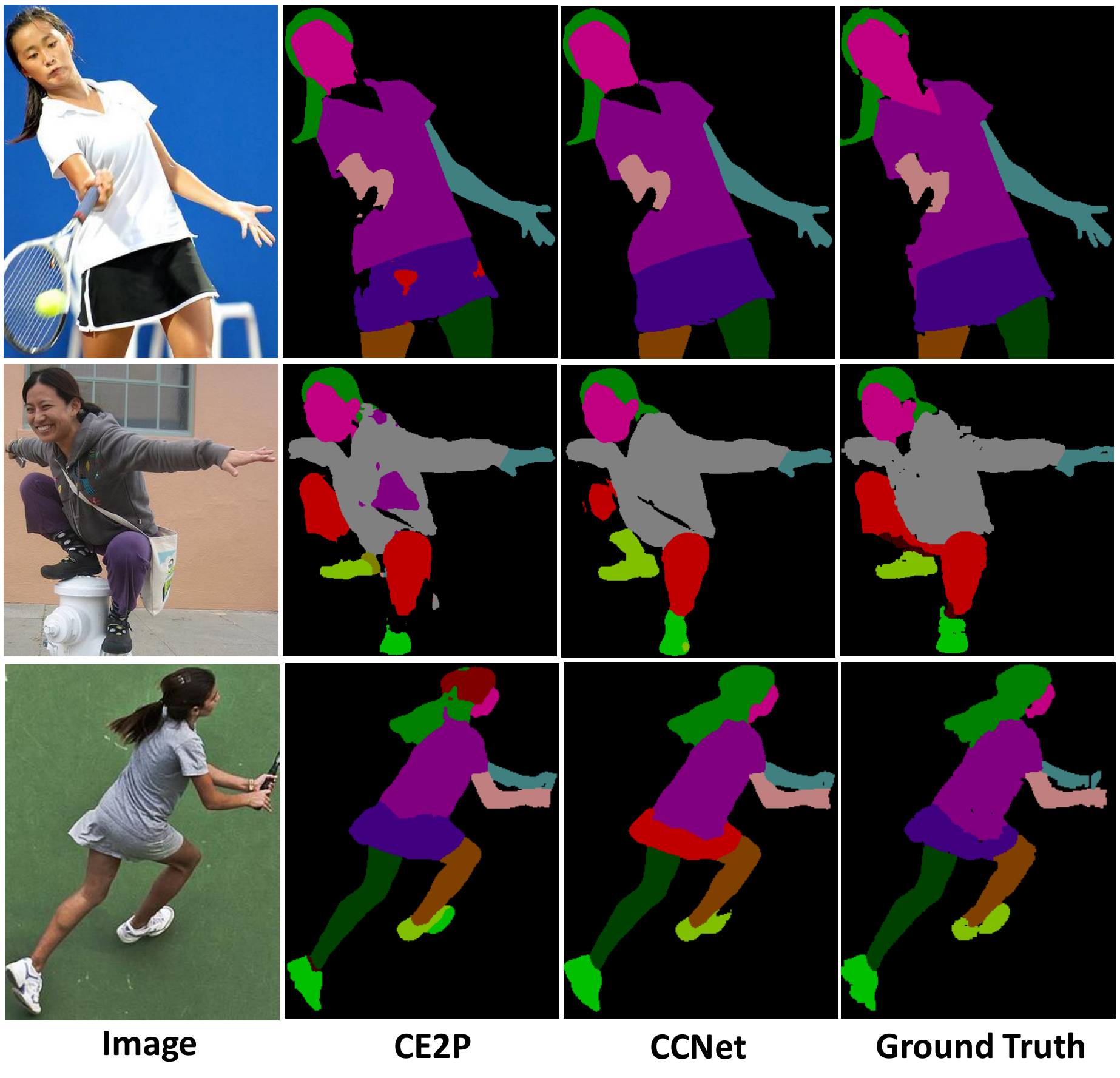}
        \caption{Visualized examples for human parsing result on LIP val set.}
        \label{fig:lip_vis}
        \vspace{-5mm}
    \end{figure}

\subsection{Experiments on LIP}

In this subsection, we conduct experiments on the LIP dataset, which is a very challenging human parsing dataset. The framework of CE2P~\cite{ruan2019devil}  is utilized, with ImageNet pre-trained ResNet-101 as bockbone and using RCCA (R=2) rather than PSP~\cite{zhao2017pyramid} as context embedding module. The category consistent loss is used to boost the performance.
The hyper-parameter setting strictly follows that in the
CE2P~\cite{ruan2019devil}. Among the approaches, Deeplab (VGG-16)~\cite{chen2014semantic}, Attention~\cite{chen2016attention} and SAN~\cite{huang2019sanet} adopt the VGG-16 as backbone and Deeplab (ResNet-101)~\cite{chen2018deeplab}, JPPNet~\cite{liang2018look}, CE2P~\cite{ruan2019devil} and CCNet adopt ResNet-101 as the backbone.
As shown in Tab.~\ref{tab:lip}, CCNet achieves the state-of-the-art performance of 55.47\%, outperforms the previous state-of-the-art methods by more than 2.3\%. This significant improvement demonstrates the effectiveness of proposed method on human parsing task. Fig.~\ref{fig:lip_vis} shows some visualized segmentation results. 
The top two rows show some successful segmentation results
It shows our method can produce accurate segmentation even for complicated poses. The third row shows a failure segmentation result where the ``skirt'' is misclassified as ``pants''. But it's difficult to recognize even for humans.

    \begin{figure*}[!t]
        \centering
        \includegraphics[width=1.0\linewidth]{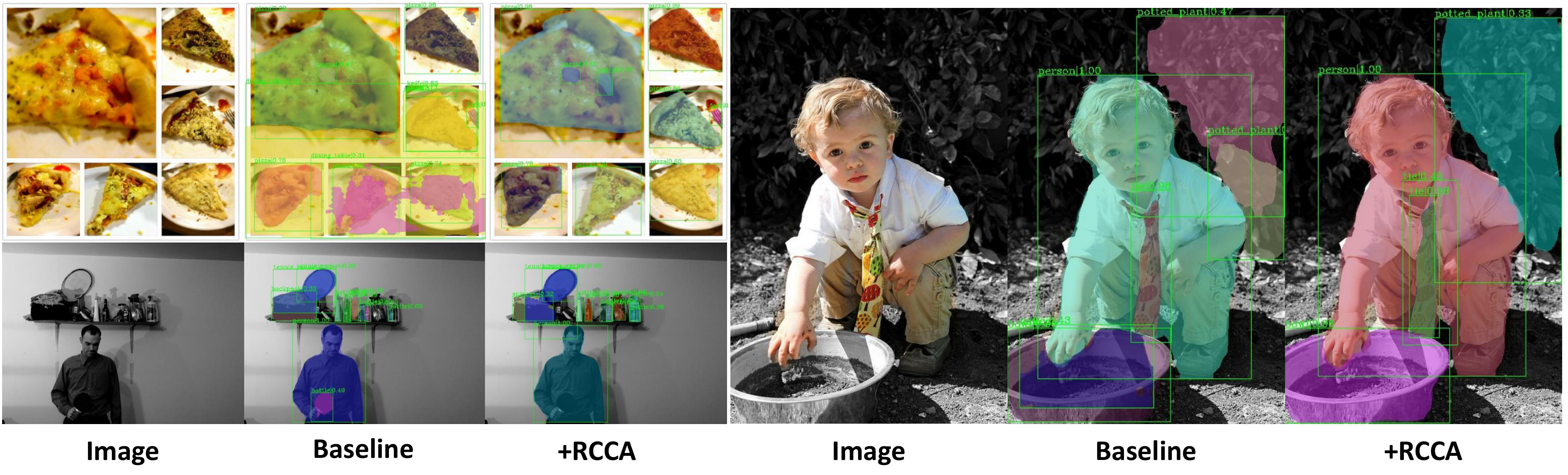}
        \caption{Visualized examples for instance segmentation result on COCO val set.}
        \label{fig:coco_vis}
    \end{figure*}
    
\subsection{Experiments on COCO}

To further demonstrate the generality of CCNet, we conduct the instance segmentation task on COCO~\cite{lin2014microsoft} using the competitive Mask R-CNN model~\cite{he2017mask} as the baseline. Following~\cite{wang2018non}, we modify
the Mask R-CNN backbone by adding the RCCA module right before the last convolutional residual block of res4. We evaluate a standard baseline of ResNet-50/101. All models are fine-tuned from ImageNet pre-training. We use the official implementation\footnote{\url{https://github.com/facebookresearch/maskrcnn-benchmark}} with end-to-end joint training whose performance is almost the same as the baseline reported in \cite{wang2018non}. 
For fair comparison, we do not use the category consistent loss in our method. We report the results in terms of box AP and mask AP in Tab.~\ref{tab:coco} on COCO. The results demonstrate that our method substantially outperforms the baseline in all metrics. Some segmentation results for comparing baseline with ``+RCCA'' are given in Fig~\ref{fig:coco_vis}. Meanwhile, the network with ``+RCCA'' also achieves the better performance than the network with one non-local block ``+NL''.

\subsection{Experiments on CamVid}

To further demonstrate the effectiveness of 3D-RCCA, we carry out the experiments on CamVid~\cite{brostow2008segmentation}, which is one of the first datasets focusing on video semantic segmentation for driving scenarios. We follow the standard protocol proposed in~\cite{badrinarayanan2017segnet} to split the dataset into 367 training, 101 validation and 233 test images. For fair comparison, we only report single-scale evaluation scores. As can be seen in Tab.~\ref{tab:camvid}, we achieve an mIoU of 79.1\%, outperforming all other methods by a large margin.

To demonstrate the effectiveness of our proposed techniques, we perform training under the same settings with the different length of input frames. We apply the CNNs on each frame for extracting features and then concatenate and reshape them to satisfy the required shape of 3D Criss-Coss Attention module. We use the $R=3$ for collecting dense spatial and temporal contextual information.
Here, to make a training sample, we try two kinds of length ($T$) of input frames. For $T=1$, we randomly sample 1 frame from a training video, donated as ``CCNet3D ($T=1$)''. For $T=5$, we sample 5 temporally ordered frames from a training video, donated as ``CCNet3D ($T=5$)''. As can be seen in Tab.~\ref{tab:camvid}, ``CCNet3D ($T=5$)'' outperforms ``CCNet3D ($T=1$)'' by 1.2\%.

    \begin{table}[!t]
        \renewcommand{\arraystretch}{1.3}
        \setlength{\tabcolsep}{1em}
        \caption{Comparison with state-of-the-arts on ADE20K (val).}
        \label{tab:ade20k}
        \centering \small
        \begin{tabular}{|l|c|c|}
            \hline
            Method & Backbone & mIOU(\%)  \\
            \hline
            RefineNet~\cite{lin2017refinenet} & ResNet-152 & 40.70\\
            SAC~\cite{zhang2017scale} & ResNet-101 & 44.30\\
            PSPNet~\cite{zhao2017pyramid} & ResNet-101 & 43.29 \\
            PSANet~\cite{zhao2018psanet} & ResNet-101 & 43.77 \\
            DSSPN~\cite{liang2018dynamic} & ResNet-101 & 43.68 \\
            UperNet~\cite{xiao2018unified} & ResNet-101 & 42.66 \\
            EncNet~\cite{zhang2018context} & ResNet-101 & 44.65 \\
            \hline
            CCNet & ResNet-101 & \textbf{45.76}\\
            \hline
        \end{tabular}
        
    \end{table}
    
        \begin{table}[!t]
        \renewcommand{\arraystretch}{1.3}
        \setlength{\tabcolsep}{0.8em}
        \caption{Comparison with state-of-the-arts on LIP (val).}
        \label{tab:lip}
        \centering \small
        \begin{tabular}{|l|c|c|c|}
            \hline
            Method & pixel acc & mean acc & mIoU  \\
            \hline
            DeepLab (VGG-16)~\cite{chen2018deeplab} &  82.66 & 51.64 & 41.64\\
            Attention~\cite{chen2016attention} & 83.43 & 54.39 & 42.92\\
            SAN~\cite{huang2019sanet} & 84.22 & 55.09 & 44.81 \\
            DeepLab (ResNet-101)~\cite{chen2018deeplab} & 84.09 & 55.63 & 44.80 \\
            JPPNet~\cite{liang2018look} & 86.39 & 62.32 & 51.37 \\
            CE2P~\cite{ruan2019devil} & 87.37 & 63.20 & 53.10 \\
            \hline
            CCNet & \textbf{88.01} & \textbf{63.91} & \textbf{55.47} \\
            \hline
        \end{tabular}
    \end{table}
    
    \begin{table}[!t]
        \renewcommand{\arraystretch}{1.3}
        \setlength{\tabcolsep}{1.2em}
        \caption{Comparison on COCO (val).}
        \label{tab:coco}
        \centering \small
        \begin{tabular}{ | l  l | c | c | }
            \hline
            \multicolumn{2}{|c|}{Method} & AP$^{box}$ & AP$^{mask}$ \\ \hline
            \multirow{3}{*}{R50} & baseline &  38.2 & 34.8 \\ 
            & +NL     &  39.0 & 35.5 \\ 
            & +RCCA     & \textbf{39.3} & \textbf{36.1} \\ \hline
            \multirow{3}{*}{R101} & baseline &  40.1 & 36.2 \\ 
            & +NL     &  40.8 & 37.1 \\ 
            & +RCCA     & \textbf{41.0} & \textbf{37.3} \\ \hline
            
        \end{tabular}
    \end{table}
    
        \begin{table}[!t]
        \renewcommand{\arraystretch}{1.3}
        \setlength{\tabcolsep}{1.2em}
        \caption{Results on the CamVid test set.}
        \label{tab:camvid}
        \centering \small
        \begin{tabular}{ | l | c | c | c | }
            \hline
            Method & Bockbone & mIoU (\%) \\ \hline
            SegNet~\cite{badrinarayanan2017segnet} & VGG16 & 60.1  \\ 
            RTA~\cite{huang2018efficient}    & VGG16 &  62.5 \\
            Dilate8~\cite{yu2015multi} & Dilate & 65.3 \\ 
            BiSeNet~\cite{yu2018bisenet} & ResNet18 & 68.7 \\ 
            PSPNet~\cite{zhao2017pyramid}  & ResNet50 & 69.1 \\ 
            DenseDecoder~\cite{bilinski2018dense} & ResNeXt101 & 70.9 \\ 
            VideoGCRF\ddag~\cite{chandra2018deep}    & ResNet101 & 75.2 \\
            \hline
            CCNet3D (T=1)~\ddag & ResNet101 & 77.9 \\
            CCNet3D (T=5)~\ddag & ResNet101 & \textbf{79.1} \\
            \hline
        \end{tabular}
        \begin{tablenotes} 
        \item \ddag ~the initialized model is pre-trained on Cityscapes.
        \end{tablenotes}
    \end{table}

\section{Conclusion and future work} \label{Conclusion}
In this paper, we have presented a Criss-Cross Network (CCNet) for deep learning based dense prediction tasks, which adaptively captures contextual information on the criss-cross path. To obtain dense contextual information, we introduce RCCA which aggregates contextual information from all pixels. The experiments demonstrate that RCCA captures full-image contextual information in less computation cost and less memory cost. Besides, to learn discriminative features, we introduce the category consistent loss. Our CCNet achieves outstanding performance consistently on several semantic segmentation datasets, \ie, Cityscapes, ADE20K, LIP, CamVid and instance segmentation dataset, \ie, COCO. The source codes of CCNet are released to facilitate related research and applications. 

\section*{Acknowledgements}
This work was in part supported by NSFC (No. 61733007 and No. 61876212), ARC DECRA DE190101315, ARC DP200100938, HUST-Horizon Computer Vision Research Center, and IBM-ILLINOIS Center
for Cognitive Computing Systems Research (C3SR) - a research collaboration as part of the IBM AI Horizons Network.


\ifCLASSOPTIONcaptionsoff
  \newpage
\fi



%
{\small
    \bibliographystyle{IEEEtran}
    \bibliography{CCNet}
}

%



\end{document}